\documentclass{article}

     \usepackage[preprint]{neurips_2020}

\usepackage[utf8]{inputenc} % allow utf-8 input
\usepackage[T1]{fontenc}    % use 8-bit T1 fonts
\usepackage{hyperref}       % hyperlinks
\usepackage{url}            % simple URL typesetting
\usepackage{booktabs}       % professional-quality tables
\usepackage{amsfonts}       % blackboard math symbols
\usepackage{nicefrac}       % compact symbols for 1/2, etc.
\usepackage{microtype}      % microtypography

\usepackage{natbib}
\usepackage{amsmath}
\usepackage{subcaption}
\usepackage{multirow}
\usepackage{pifont}% http://ctan.org/pkg/pifont
\usepackage{array}
\usepackage{algpseudocode,algorithm,algorithmicx}  

\usepackage{graphicx} %package to manage image

\graphicspath{ {images/} }

\title{Impression Space from Deep Template Network}

\author{%
  Gongfan Fang \\
  Zhejiang University \\
  \texttt{fgf@zju.edu.cn} \\
   \And
   Xinchao Wang \\
   Stevens Institute of Technology \\
   \texttt{xinchao.wang@stevens.edu} \\
   \AND
   Haofei Zhang \\
   Zhejiang University \\
   \texttt{haofeizhang@zju.edu.cn} \\
   \And
   Jie Song \\
   Zhejiang University \\
   \texttt{sjie@zju.edu.cn} \\
   \And
   Mingli Song \\
   Zhejiang University \\
   \texttt{brooksong@zju.edu.cn} \\
}

\begin{document}

\maketitle

\begin{abstract}

It is an innate ability for humans to imagine something only according to their impression, without having to memorize all the details of what they have seen. In this work, we would like to demonstrate that a trained convolutional neural network also has the capability to ``remember'' its input images. To achieve this, we propose a simple but powerful framework to establish an {\emph{Impression Space}} upon an off-the-shelf pretrained network. This network is referred to as the {\emph{Template Network}} because its filters will be used as templates to reconstruct images from the impression. In our framework, the impression space and image space are bridged by a layer-wise encoding and iterative decoding process. It turns out that the impression space indeed captures the salient features from images, and it can be directly applied to tasks such as unpaired image translation and image synthesis through impression matching without further network training. Furthermore, the impression naturally constructs a high-level common space for different data. Based on this, we propose a mechanism to model the data relations inside the impression space, which is able to reveal the feature similarity between images. Our code will be released. 

\end{abstract}

\section{Introduction}
In the past few years, deep learning has become a mainstream paradigm for tackling computer vision problems. The classification task, as the pioneer to bring up deep learning techniques \citep{krizhevsky2012imagenet}, plays a significant role in the field of computer vision. For example, pretraining is a widely used technique derived from classification. A pretrained deep neural network usually has powerful representation ability \citep{azizpour2015factors} and can provide huge benefits for solving various sophisticated tasks such as semantic segmentation \citep{chen2017deeplab} and object detection \citep{ren2015faster}. 

Image generation is another important topic in the field of computer vision. Many techniques, including Generative Adversarial Networks (GANs) \citep{goodfellow2014generative} and Autoencoders \citep{kingma2013auto} provide us with powerful framework for solving various generation problems such as image synthesis \citep{brock2018large}, image-to-image translation \citep{zhu2017unpaired} and Attribute Manipulation \citep{shen2017learning}. Recently, some classifier-based generation methods have also received widespread attention and achieved impressive results by classifier inversion \citep{lopes2017data,santurkar2019image,yin2019dreaming}. However, in the image generation literature, a prominent problem of the existing frameworks is that their generation ability is usually limited to a specific domain, i.e., the domain of training data. To synthesize a new category, abundant training samples are required, and burdensome network training is usually inevitable, which is very expensive. So, is there a universal image synthesizer that can synthesize arbitrary images and does not require further network training? In this work, we would like to show that all we need is just a pretrained classifier.

In this work, we propose the \emph{Impression Space}, which is established on an off-the-shelf classifier through a simple encoding and decoding paradigm. This space actually constitutes a powerful feature dictionary to describe images and can be utilized to tackle many sophisticated tasks, including one-shot image synthesis, unpaired image-to-image translation, and feature similarity estimation. In our framework, we will refer to the pretrained classifier as the \emph{Template Network} because its filters will be used as templates for image synthesizing. More strikingly, our framework is universal and can be generalized to any given images, even if they are not from the domain of the original training data. Our contributions can be summarized as follows:

\begin{itemize}
  \item We propose a framework for tackling sophisticated tasks such as one-shot image synthesis, image-to-image translation, and feature similarity estimation \emph{only} with a pretrained classifier.
  \item Our framework provides an alternative algorithm for GANs and is very simple to practice. 

\end{itemize}

\begin{figure*}[t]
  \centering
  \includegraphics[width=13.5cm]{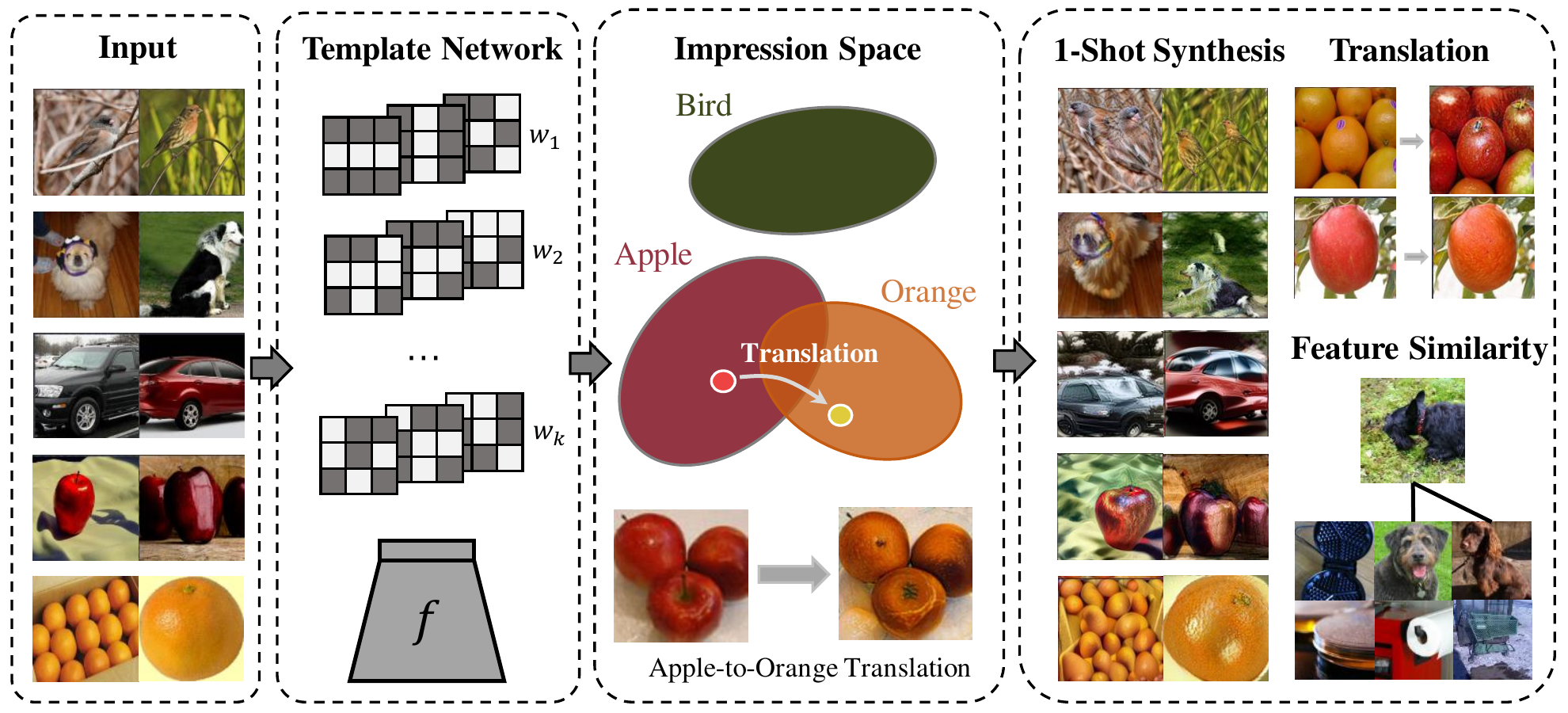}
  \caption{We propose impression space to tackle various computer vision tasks. First, the input images are fed to a pretrained template network and embedded into the impression space. Then we can synthesize images from the impression or translate images into another domain (e.g., apple-to-orange translation) through a simple impression matching process. Our framework also provides powerful tools to estimate feature similarity and can be applied to data selection.}
  \label{fig:framework}
  \vspace{-5mm}
\end{figure*}

\section{Related Works}

\subsection{Image Synthesizing and Translation.} Synthesizing realistic and natural images is still a challenging problem in the field of computer vision. In the past few years, many techniques such as Generative Adversarial Networks (GAN)  \citep{goodfellow2014generative,brock2018large} and Autoencoders \citep{kingma2013auto} have achieved impressive advancements and became a popular paradigm for image generation. In addition to these mainstream frameworks, another type of generation methods based on pre-trained classifiers have also received widespread attention in the recent years \citep{lopes2017data,santurkar2019image,yin2019dreaming}. Lopes et al. introduce a data-free knowledge distillation framework \citep{lopes2017data,hinton2015distilling} by reconstructing training data from the layer-wise statistics stored during training. Yin et al. introduce the DeepInversion \citep{yin2019dreaming} method to invert a classifier and recover its training data by matching the mean and variance statistics stored in Batch Normalization \citep{ioffe2015batch}. Santurkar et al. demonstrate that a robust classifier can provide powerful toolkits for many visual tasks, including images synthesizing \citep{santurkar2019image}. Earlier related work can be traced back to the previous studies on the interpretability of deep neural networks \citep{mordvintsev2015inceptionism,mahendran2015understanding,simonyan2013deep}, which are trying to inverse certain neurons to understand the network. However, in our work, we will focus on a more challenging problem and try to build a universal synthesizer for any given images. 

Image-to-image translation is another crucial problem in computer vision, which involves understanding the semantic from paired or unpaired images \citep{zhu2017unpaired}. In the past few years, the GAN framework has become the most popular method for solving image-to-image translation tasks and achieved impressive results on several datasets \citep{zhu2017unpaired,isola2017image}. However, most of the GAN-based methods cannot provide a universal translator, which means that for each dataset, we have to collect a large amount of training data and independently train a network to achieve image translation. In contrast, we will show that our framework can provide a universal translator to tackle this problem. 

\subsection{Transferability and Feature Similarity.} Transfering a trained network from one task to another is a widely used technique in deep learning  \citep{chen2017deeplab,ren2015faster}. The features learned by a deep neural network can be used as generic image representations to tackle various tasks  \citep{sharif2014cnn}. However, the effect of transfer learning largely depends on the gap between data. Many works have been proposed to quantify the difference between data \citep{torralba2011unbiased,tommasi2017deeper}, which is useful for image retrieval and transfer learning. Our work provides a new perspective of data difference measurement and verifies it with experiments on knowledge transfer.      

\section{Methods}

Given only one image, we will show that an off-the-shelf classifier has the capability to capture the salient features from it and synthesize new images, which can be viewed as one-shot image generation. Besides, such a classifier can also be utilized to tackle more sophisticated tasks, including unpaired image-to-image translation \citep{zhu2017unpaired} and image retrieval \citep{gordo2016deep}. In this section, we propose the $Impression Space$, which can be directly established upon a pretrained classifier without any further network training. 

\subsection{Impression Space}
The key idea of the impression space in simple: we can describe an image with a predefined feature dictionary. This idea is widely used in many traditional vision methods, such as the Bag-of-Words model \citep{zhang2010understanding}. However, building an appropriate feature dictionary manually is usually intractable due to the diversity of natural images. In this work, we demonstrate that the convolutional filters learned by a pretrained classifier naturally provides a powerful feature dictionary. Our framework is illustrated in Figure \ref{fig:framework}. Given an image $x$ and a bunch of convolutional filters $w=[w_1, w_2, ..., w_K]$ from a pretrained classifier, we encode the images into a \emph{Impression Code} $z$ by calculating a channel-wise mean statistics as follows:
\begin{equation}
  z = \left[ \mu(r_1), \mu(r_2), ..., \mu(r_K) \right]  \label{eqn:enc}
\end{equation}

where $r_i$ is the intermediate representation produced by the filter $w_i$ and $\mu(\cdot)$ takes the average statistic on all elements in $r_i$. Note that a convolutional operation can be regarded as a series of inner products, the mean statistic actually reveals the response of a certain pattern in the given image. This coding paradigm establishes a space, which is referred to as \emph{Impression Space} in this work. In the following sections, we would like to shed some light on the impression space and discuss how it can be leveraged to tackle various generation tasks.

\subsection{Decoding Impression Code for Image Synthesis}

Image synthesis is one of the most basic ability of impression space. Given a images $\hat{x}$, we first encode it into its impression code $\hat{z}$. Then synthsizing images from the domain of $\hat{x}$ is simple: we randomly initialize some noisy images $x$ and then update them to match their impression code $z$ to the $\hat{z}$, which can be achieved by solving an optimization problem:
\begin{equation}
  x = \mathop{\arg \min}_{x} \frac{1}{2} \left\|z - \hat{z}\right\|^2
  \label{eqn:dec}
\end{equation}

\textbf{Connection to LSGAN.} In LSGANs \citep{mao2017least}, the min-max game is setup between a generator $G$ and a Discriminator $D$ as follows:
\begin{equation}
  \begin{split}
    &\mathop{\min}_{D} V(D) = \frac{1}{2} \mathbb{E}_{\hat{x}\sim p_{data}(\hat{x})}\left[(D(\hat{x})-b)^2\right] + \frac{1}{2} \mathbb{E}_{z\sim p_{z}(z)}\left[(D(G(z))-a)^2\right] \\
    &\mathop{\min}_{G} V(G) = \frac{1}{2} \mathbb{E}_{z\sim p_{z}(z)}\left[(D(G(z))-c)^2\right]
  \end{split}
\end{equation}

The discriminator $D$ can be viewed as a classifier, which tries to distinguish the generated samples from the real ones according to the response of certain patterns. Analogously, in our framework, the response of an intermediate feature map $r_i$ also indicates whether certain features exist in the images. Hence, the filter $w_i$, combined with its previous layers, can be regarded as a trained and fixed discriminator, denoted as $D_i^*$. This analogy is reasonable because there is no essential difference between a classifier and a discriminator as their architectures and loss functions are similar. For each fixed $D_i^*$, we can set $a_i=c_i=\hat{z_i}$ and $b=z_i$. Then Equation \ref{eqn:dec} can be rewritten in a similar form as the problem from LSGANs:
\begin{equation}
    x = \mathop{\min}_{x} V(x) = \frac{1}{2} \sum_i (D_i^*(x)-c_i)^2
\end{equation}
In this problem, $D^*=[D_1^*, D_2^*, ..., D_K^*]$ can be viewed as an ensemble of a set of fixed discriminators. The main difference between this process and LSGANs is that our discriminator is not obtained from training. Instead, we utilize the filters from a pretrained network as templates to implicitly distinguish real samples from synthesized ones. On the other hand, if we can construct an ideal classifier with infinitely rich templates, then it is accessible to replace the discriminator in GAN training with templates appropriately selected from the ideal classifier. 

According to the above analysis, the quality of the synthesized image in our framework is mainly determined by the diversity of $D*$. In other words, an obvious conclusion is that those classifier networks with diverse filters usually have stronger synthesis capabilities. Besides, the synthesis ability is still of a pretrained network can be generalized to any given images, leading to the universal image synthesis. When we only have one input image, this training process is trying to place those template features onto the image, leading to the one-shot image generation.

\textbf{Improve Image Quality with Variance.} In previous works, variance and covariance are proposed to be useful for inverting a classifier \citep{yin2019dreaming,lopes2017data}. However, we find that these statistics do not change the content or semantic of the synthesized images. In contrast, introducing variance can improve the diversity of features in spatial so as to improve the image quality. So we can integrate the variance into our impression code to get a better image quality as follows:
\begin{equation}
  z = \left[ \mu(r_1), \sigma^2(r_1), ..., \mu(r_K), \sigma^2(r_K) \right] \label{eqn:enc}
\end{equation}

\subsection{Capture Salient Features by Averaging} \label{sec:additive}

Many vision tasks such unpaired image-to-image translation require the network to catch some salient features from images. For example, to translate horses to zebras \citep{zhu2017unpaired}, the network should at least learn how to distinguish the horse from background \citep{santurkar2019image}. In this work, we demonstrate that the impression space actually provides such ability to extract the salient features from images. Given a set of images $X=[x^1, x^2, ..., x^N]$, the salient features can be catched by applying Equation \ref{eqn:enc} on the whole dataset. Specifically, we calculate the mean and variance for across all the images to get a ensembled impression code $z^*$ as follows:
\begin{equation}
  z^* = \left[ \mu(r_1^1, ..., r_K^1), \sigma^2(r_1^1, ..., r_K^1), ..., \mu(r_1^N, ..., r_K^N), \sigma^2(r_1^N, ..., r_K^N) \right] \label{eqn:enc_avg}
\end{equation}

\vspace{-2mm}
where $r_i^n$ denotes the $i$-th intermediate representation from $n$-th samples $x_n$. $\mu(\cdot)$ and  $\sigma^2(\cdot)$ calculate the mean and variance across all input examples in a similar way as  \citep{ioffe2015batch}. The motivation behind this approach is the salient features of images will appear repeatedly and will still maintain a high activation after averaging. In contrast, the negligible features from background objects will be erased due to fewer occurances. The above mentioned ensembled $z^*$ provides a powerful tool to tackle image manipulation problems, such as unpaired image-to-image translation, which can be achieved by solving a similar problem as image synthesis. Given as set of source $x_s$ and target images $x_t$, we first encode the target images into the ensembled impression code $z_t^*$. And then apply an iterative optimization on the source images to solve the following problems:
\begin{equation}
  x = \mathop{\arg \min}_{x} \left\|z - z_t^*\right\|^2 + \lambda \left\|x - x_s\right\|^2 \label{eqn:translation}
\end{equation} 
where $x$ is initialized from $x_s$ and $z$ is its impression code. The impression matching is done under the  of restrained by a content regularization with a balance term $\lambda$. 
\begin{algorithm}[t]
  \caption{Image-to-Image Translation in Impression Space}\label{algorithm:impression_space}
  
  \hspace*{\algorithmicindent} 
    \textbf{Input:} \hspace*{0.2cm} $\mathbf{x_s}$: source images to be translated; \\
    \hspace*{1.9cm} $\mathbf{x_t}$: target images; \\
    \hspace*{1.9cm}  $f$: a pretrained classifier.
  \\
  \hspace*{\algorithmicindent} \textbf{Output:} $\mathbf{x}$: the translated images.
  \begin{algorithmic}[1]
  \Procedure{TRANSLATION}{$x_s, x_t, f$} % \Comment{The}
      \State $ x \gets x_s$ 
      \Comment{Initialize from the source images}

      \State $z_t \gets Enc(f, x_t)$ 
      \Comment{Obtain the target impression by Eqn. \ref{eqn:enc_avg}}

      \For{$i \gets 1 \textrm{ to } k$}
              \State $z \gets Enc(f, x)$ 
              \Comment{Obtain the image impression by Eqn. \ref{eqn:enc_avg} or \ref{eqn:enc} }

              \State 
                  $ L \gets || z - z_t ||^2 + \lambda ||x - x_s||^2$ 
              \State 
                  $\mathbf{x} \gets \mathbf{x} - \eta \nabla_{\mathbf{x}_i} L$
                  \Comment{Update images to match impression.}
      \EndFor
      \State \textbf{return} $x$
  \EndProcedure
  \end{algorithmic}
\end{algorithm}
\subsection{Impresion Distance}

All the above-mentioned algorithms are based on a very simple metric, that is, the distance inside the impression space. In this section, we will show its applications in feature similarity estimation. Feature similarity plays an important role in deep learning \citep{xu2019positive,coleman2019selection}. Often we want to transfer the knowledge from a trained network to another one \citep{hinton2015distilling,tian2019contrastive}, but only have a few labeled data \citep{xu2019positive}. A common solution to tackle this problem is to use some cheap and unlabeled data \citep{blum1998combining}. To select suitable unlabeled samples, we need to evaluate the similarity between unlabeled images and labeled ones. As aforementioned, the impression code $z$ utilizes the trained filter from the template network to describe images in the feature-level. Such a description is invariant to some factors such as spatial position and naturally provides a metric for the feature similarity estimation. We will show this in the experiment part.

\section{Experiments}

In the this sections, we demonstrate our results on three task: image synthesis\citep{santurkar2019image}, unpaired image-to-image translation\citep{zhu2017unpaired} and feature similarity estimation\citep{xu2019positive}. For image synthesis and translation, we adopt an off-the-shelf ResNet-50\citep{he2016deep} pretrained on ImageNet\citep{deng2009imagenet} as the templates classifier and use Adam as our optimizer. Detailed experimental settings can be found in supplementary materials.

\subsection{Synthesize Anything You Like.}  

\begin{figure*}[t]
  \centering
  \includegraphics[width=13.5cm]{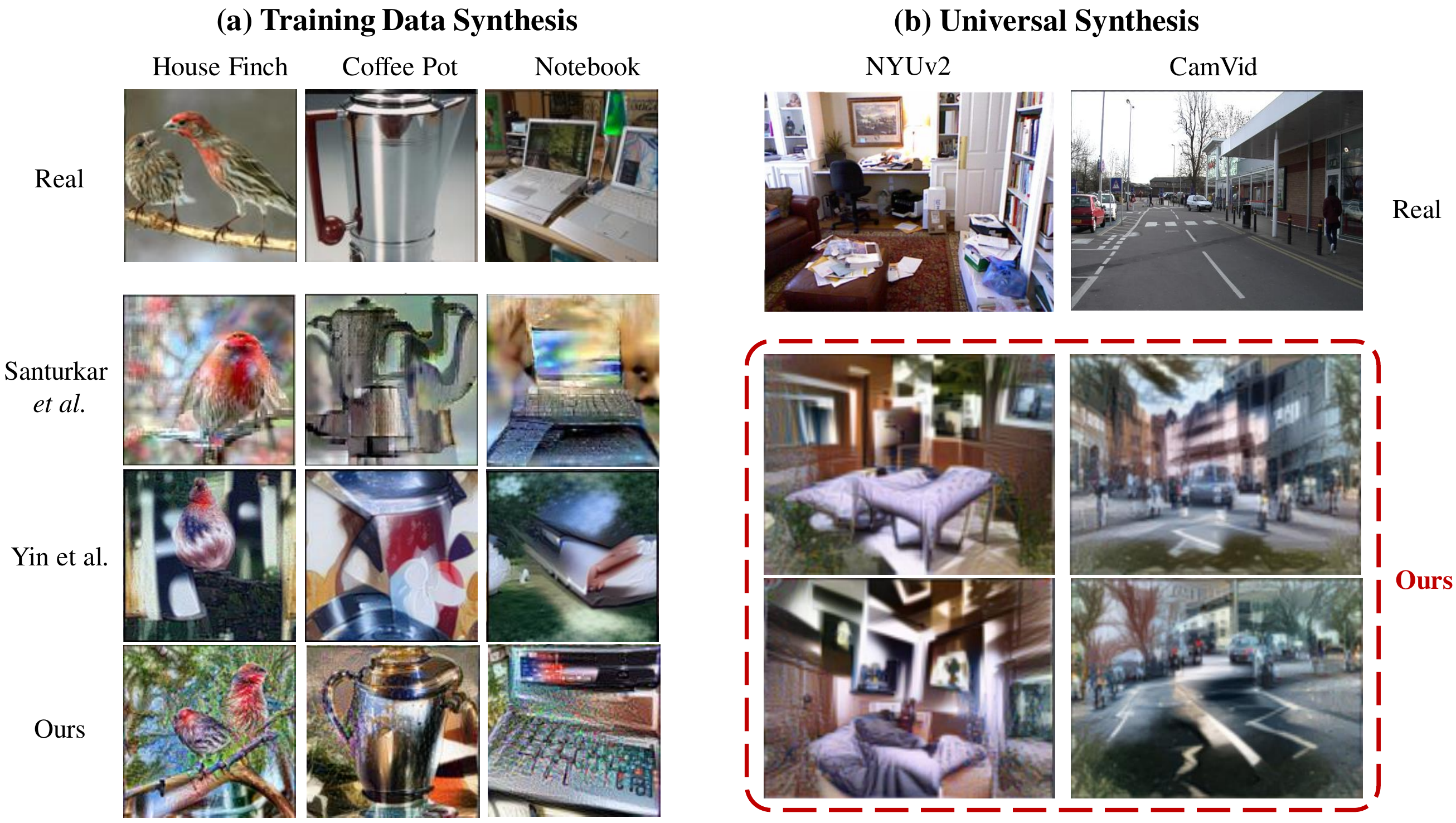}
  \caption{Synthesized images from different classifier-base methods. \textbf{left}: synthesize images from original training data (i.e. ImageNet). \textbf{right}: synthesize any given images with the proposed impression space. Please refer to the supplementary materials for more results.} 
  \label{fig:synthesis}
  \vspace{-2mm}
\end{figure*}

In this section, we will show that a pretrained classifier has the ability to synthesize any image. This can be easily achieved by encoding a set of images into the impression and then decoding it back to the image space. We compare our method with the previous classifier-based methods proposed in \citep{santurkar2019image,yin2019dreaming} and illustrate the synthesized images in Figure \ref{fig:synthesis}. These baseline methods utilize the class score to control the output and can not generate objects outside its training data. In contrast, our method is able to synthesize anything by showing images to the network. Figure \ref{fig:synthesis} (b) displays the results obtained from CamVid\citep{brostow2008segmentation} and NYUv2\citep{Silberman:ECCV12}. However, there is still a shortcoming in our method, which is that it requires real-world images for synthesis while these baseline methods are data-free \citep{yin2019dreaming}.

\begin{figure*}[t]
  \centering
  \includegraphics[width=13.7cm]{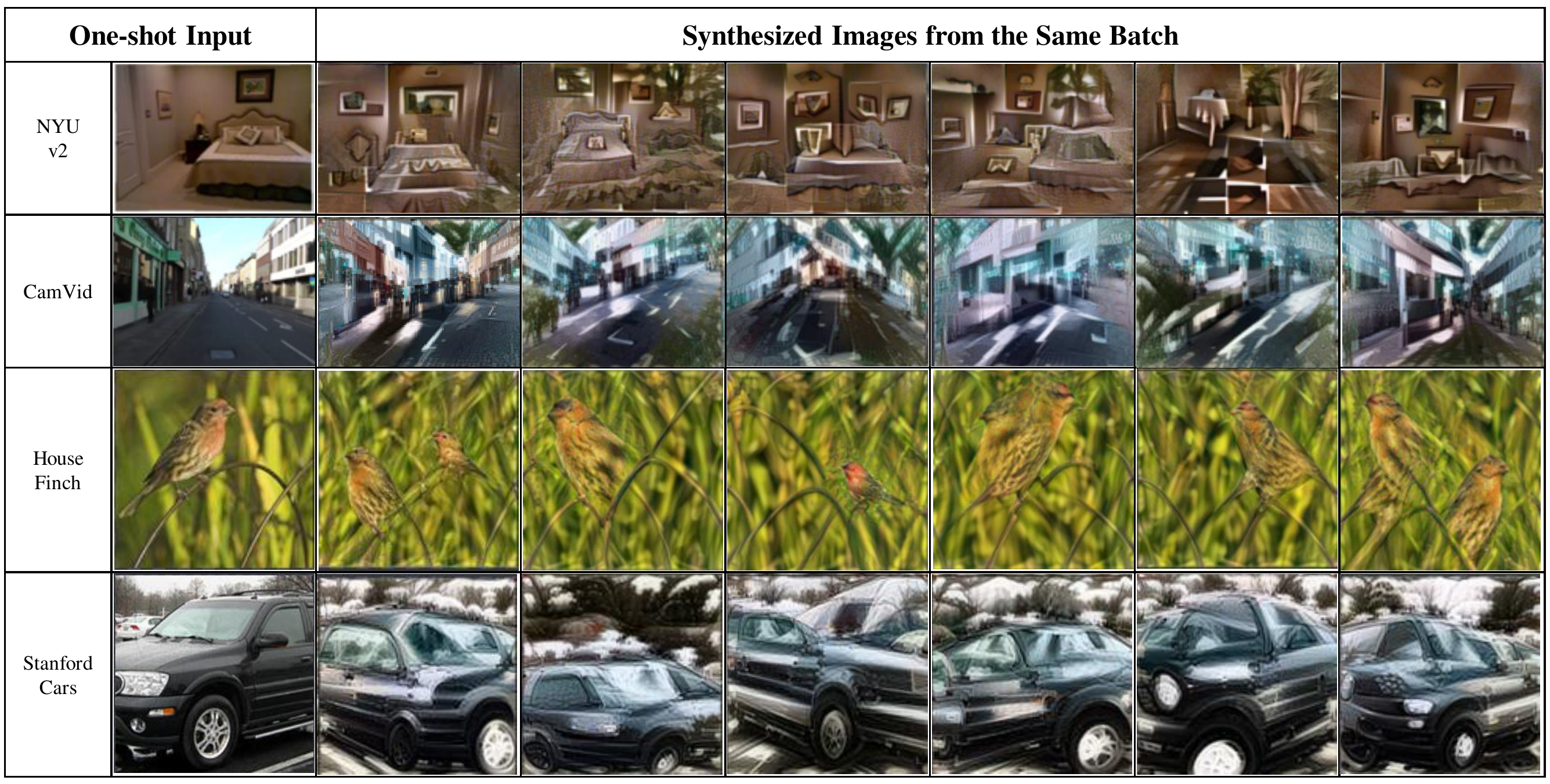}
  \caption{Image diversity in one-shot synthesis. All the images are randomly initialized and come from the same batch.} 
  \label{fig:oneshot}
  \vspace{-5mm}
\end{figure*}

\subsection{Diversity in One-shot Image Synthesis}  
In this part, we will provide more details about one-shot image synthesis. To generate images from a new category, most GAN-based algorithms require abundant training images, which is very expensive. Recently, some few-shot algorithms \citep{clouatre2019figr} are proposed under the GAN framework. However, it is still difficult to extend those methods to real-world images. Now, we demonstrate that the template filters in a pretrained classifier are sufficient enough to achieve one-shot image generation. Given only one image, the goal of one-shot image generation is to synthesize some different images but with similar semantic as the given one, as shown in \ref{fig:oneshot}. We display the input image in the first column and randomly pick six synthesized images for visualization. These synthesized images indeed preserve similar semantics as the input image and are diverse in details. The Inception Score (IS) and Fréchet Inception Distance (FID) of our method can be found in Table \ref{tbl:IS}. Note that our goal is to build a universal synthesizer for any input images rather than achieve high IS or FID on a certain dataset.

\newcommand{\cmark}{\ding{51}}%
\newcommand{\xmark}{\ding{55}}%
\begin{table}[t]
  \centering 
  \caption{The Inception Score (IS) and Fréchet Inception Distance (FID) from different methods on ImageNet. The item "Training" indicates whether network training is required and the item "Universal" indicates whether the method can be used to synthesize any given image. Some FID results are not proposed in the original paper.} \label{tbl:IS}
  \begin{tabular*}{\textwidth}{l @{\extracolsep{\fill}} c c c c c c}
    \hline\hline
    %Method & Training Data & BigGAN & WGAN-GP & DeepInversion & Robust Classifier & Ours \\
    %\hline
    %IS & 331.9 & 233.1 & 11.6 & 259.0 & 28.78 \\
    Method & GAN & Training & Universal & Resolution & IS & FID\\
    \hline
    Training Data     &    -    &  -     & -      & 224 & 331.9 &  -\\
    BigGAN            & \cmark  & \cmark & -      & 256 & 232.5 & 8.1 \\
    WGAN-GP           & \cmark  & \cmark & -      & 128 & 11.6  &  -\\
    SAGAN             & \cmark  & \cmark & -      & 128 & 52.5  & 18.7 \\
    SNGAN             & \cmark  & \cmark & -      & 128 & 35.3  & - \\
    Robust Classifier &  -      & \cmark & -      & 224 & 259.0 & 36.0 \\
    DeepInversion     &  -      &  -     & -      & 224 & 60.6  & - \\
    \hline
    Ours-Average      &  -      &  -     & \cmark & 224 & 28.9  & 29.3 \\
    Ours-OneShot      &  -      &  -     & \cmark & 224 & 30.5  & 28.0 \\
    \hline\hline                
  \end{tabular*}
  \vspace{-3mm}
\end{table}

\subsection{Unpaired Image-to-Image Translation}

\begin{figure*}[t]
  \centering
  \includegraphics[width=13.7cm]{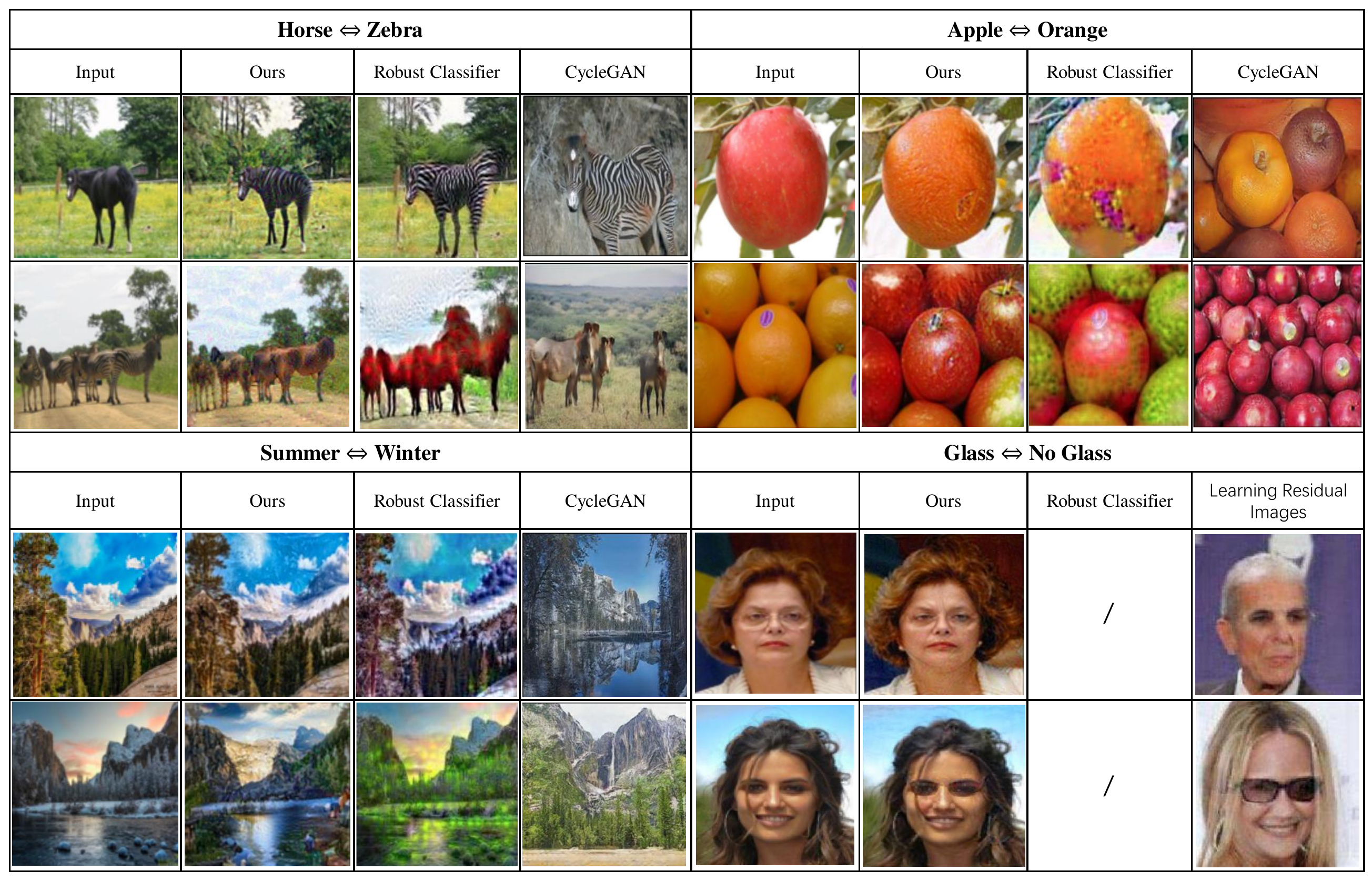}
  \caption{Unpaired image-to-image translation. Our method achieves a better translation results  than the robust classifier-based algorithm  \citep{santurkar2019image} on all datasets except the "Sunmmer to Winter" translation.} 
  \label{fig:translation}
  \vspace{-2mm}
\end{figure*}

Image translation, as another significant problem in generation tasks, has received extensive attention from researchers \citep{zhu2017unpaired,isola2017image}. Our framework treats this task as feature translation, which is very straightforward in the impression space: given a set of target images, we encode them into one impression code using Equation \ref{eqn:enc_avg}, which captures the salient features from data. Then we update the source images with a content regularization and pull the impression code to the target one. Figure \ref{fig:translation} shows our results on four different datasets. Our method performs better than the previous classifier-based method proposed by Santurkar \citep{santurkar2019image} on almost all translation directions, except in "Summer-to-Winter." Compared with GAN-based methods, our method is slightly worse in quality. However, our framework work only requires one pre-trained classifier to tackle all these translation tasks, and no network training is required. These results demonstrate that the salient features in images can be effectively captured through the ensembled impression code. Besides, we find that it is impossible to achieve one-shot image translation without prior knowledge because the network can not figure out what is the most important object from a single image.

\subsection{Impression Distance for Feature Similarity}

\begin{figure*}[t]
  \centering
  \includegraphics[width=13.7cm]{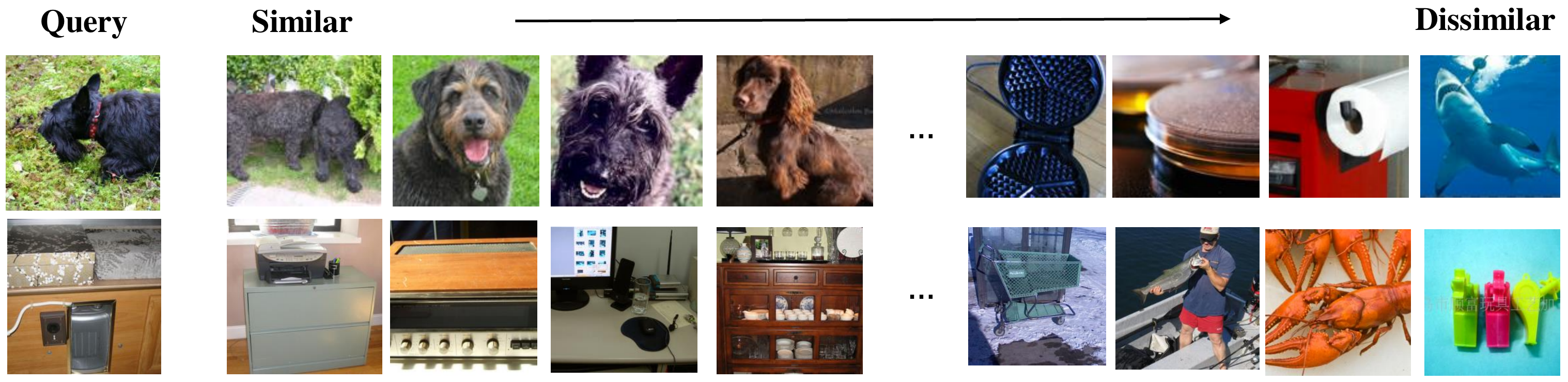}
  \caption{Feature similarity measured in Impression Space. Query images come from the training set of ImageNet. We measure the impression distance for samples from the validation set and rank them according to the similarity.} 
  \label{fig:retrieval}
  \vspace{-2mm}
\end{figure*}

In the field of image retrieval, most previous works mainly focus on content-based retrieval image \citep{cao2020unifying,radenovic2018revisiting}. Here we focus on a similar but slightly different problem, that is, data selection \citep{coleman2019selection,xu2019positive}. The goal of data selection is to choose appropriate samples from a massive labeled or unlabeled data for learning. In order to verify the effectiveness of impression distance in data selection, we conducted some qualitative and quantitative experiments, respectively. In the qualitative experiment, we perform image retrieval on ImageNet using the impression distance from a ResNet-50, as illustrated in Figure \ref{fig:retrieval}. In the quantitative experiment, we compare our methods to the Positive-Unlabeled compression method\citep{xu2019positive} on CIFAR-10\citep{krizhevsky2009learning}. Following their problem setting, we use the training set of ImageNet as the unlabeled data and use $n_l$ labeled ones from CIFAR-10 for data selection. The selected data is used to compress a trained ResNet-34 teacher to a ResNet-18 student. We utilize the pretrained teacher as our template model and encode all labeled data as an ensembled impression code by applying \ref{eqn:enc_avg}. The compression results are shown in Table \ref{tbl:retrieval}. For different $n_l$, we select $n_t = 50,000$ unlabeled samples which is close to the labeled data in impression space and resize to $32\times 32$ for training. Our method achieves stable results on all $n_l$ settings.

\begin{table}[t]
  \caption{Results of model compression on CIFAR-10 dataset.  $n_l$ and $n_t$ indicates the number of labeled data from CIFAR-10 and selected data from unlabeled ImageNet.}
  \centering
  \begin{tabular*}{\textwidth}{l @{\extracolsep{\fill}} l l l c c c}
    \hline\hline
    Method & $n_l$ & $n_t$ & Data source & FLOPs &  $\#$params & Acc($\%$)\\
    \hline
    Teacher   & - & 50,000 & Original Data & 1.16G & 21M & 95.61\\
    KD  & - & 50,000 & Original Data & 557M & 11M & 94.40\\
    Oracle    & - & 269,427 & Manually Selected Data & 557M & 11M & 93.44\\
    Random    & - & 50,000 & Randomly Selected Data & 557M & 11M & 87.02 \\
    \hline%\cdashline{1-7}[0.8pt/2pt]
    \multirow{3}{*}{PU}   & 100 & 110,608 & \multirow{3}{*}{PU data}   & \multirow{3}{*}{557M}   & \multirow{3}{*}{11M}   &  93.75\\
    & 50  & 94,803 &  &  &  & 93.02 \\
    & 20  & 74,663 &  &  &  & 92.23 \\
    \hline
    \multirow{4}{*}{Ours}   
    & 100 & 50,000 & \multirow{4}{*}{Retrieved Data}   & \multirow{4}{*}{557M}   & \multirow{4}{*}{11M}   & \textbf{93.79}\\
    & 50  & 50,000 & & & & 93.71 \\
    & 20  & 50,000 & & & & 93.56 \\
    & 5   & 50,000 & & & & 92.43 \\
    \hline
    \hline
  \end{tabular*} \label{tbl:retrieval}
\end{table}

\section{Conclusion}

In this work, we propose a simple but powerful framework, which is able to tackle many sophisticated tasks such as image synthesis, unpaired image-to-image translation, and feature similarity estimation only with a pretrained classifier. Our results show that it is feasible to build a universal image synthesizer with the template filters from a pretrained classifier. Our framework provides an alternative algorithm for GANs, especially when the training data is scarce or expensive to obtain. Furthermore, we also make some explorations on the relationship between our classifier-based algorithm and the GAN framework and provide an explanation for the image generation from the perspective of feature extraction. 

\clearpage

\section{Statement of Broader Impact}

The main purpose of this work is to explore the capabilities of a pretrained classifier in generative tasks, such as image synthesis and translation. Our work shows that it is feasible to construct a universal synthesizer only with one classifier, and we believe that this technique can effectively reduce the cost of solving generation tasks. However, 
every coin has two sides. The possibility of misuse of generative technology also exists, especially when the training cost becomes reduced.

{\small
 \bibliographystyle{apalike}
 \bibliography{citation}
}

\clearpage

\textbf{\Large Supplementary Material}

\renewcommand*{\thesection}{\Alph{section}}
\renewcommand*{\thesubsection}{\thesection.\arabic{subsection}}

This supplementary material is organized as three parts: Section \ref{sec:exp_setting} provides detailed settings for each experiment from the main paper; Section \ref{sec:relation} discusses the difference between our methods and some related methods; Section \ref{sec:visualization} provides more visualization results and analysis.

\section{Experimental Settings} \label{sec:exp_setting}

\subsection{Image Synthesis} We adopt an ImageNet pretrained ResNet-50 classifier as the template network, and no further network work is required. In One-shot generation, we feed the center cropped input image into the classifier and obtain the impression code by the mean-variance encoding process. During decoding, we initialize a set of noisy images with the batch size of 32 and update the noisy images to match their impression code to the given one for 2000 iterations with Adam optimizer. We set the learning of Adam to 0.1, and the betas to [0.5, 0.9]. During optimization, we apply random cropping ($224 \times 224$) and random flipping (probability=0.5) to the inputs, and we find these augmentations are beneficial to the image quality.

\subsection{Image-to-Image Translation} Image-to-Image Translation can be achieved similarly as image synthesis. We also adopt an off-the-shelf ResBet-50 classifier to tackle this task using the same optimization and augmentations settings as in image synthesis. The coefficient $\lambda$ for content regularization is listed in Table \ref{tbl:lambda}. 

\begin{table}[h]
  \centering 
  \vspace{-3mm}
  \caption{The coefficient $\lambda$ of content regularization for different image-to-image translation tasks.} \label{tbl:lambda}
  \begin{tabular}{c | c c c c}
  \hline\hline
  Task (A $\leftrightarrow$ B) & Apple $\leftrightarrow$ Orange & Horse $\leftrightarrow$ Zebra & Summer $\leftrightarrow$ Winter & Glass $\leftrightarrow$ No Glass \\
  $\lambda_{A2B}$ & 5e-5 & 5e-5 & 8e-6 & 5e-5 \\
  $\lambda_{B2A}$ & 1e-5 & 2e-5 & 8e-6 & 2e-5\\
  \hline
  \end{tabular}
\end{table}

\subsection{Feature Similarity and Image Retrieval}

In this experiment, our goal is to compress a trained ResNet-34 model to a ResNet-18 model through data selection. We use the final accuracy to prove that our method can be used to measure the feature similarity. Following \citep{xu2019positive}, we randomly pick $n_l$ labeled images from CIFAR-10 and use the ImageNet as the unlabeled dataset. First, we utilize the trained teacher as the template network and ensemble all labeled images into one impression code. Then for each image in ImageNet, we encode it into the impression space to measure its distance to the ensembled one. According to the impression distance, we select the first $n_t$ unlabeled samples as our training data. We use the same settings as \citep{xu2019positive} except that the standard Knowledge Distillation \citep{hinton2015distilling} is applied for model compression. We use SGD with the learning rate of 0.1 to update the student model, and the learning rate is decayed by 0.1 at 50, 100, and 150 epochs. We update the student model for 200 epochs and report the best accuracy. 

\section{Relation to Previous Works} \label{sec:relation}

\textbf{Generative Adversarial Networks (GANs).} Currently, GAN is one of the most popular frameworks for solving generation problems. Our framework aims at providing an alternative algorithm for GANs, especially when training data is scarce or very expensive to obtain. Our method only requires an off-the-shelf classifier and can synthesize any given image without further network training. Besides, we think that there is a certain interconnectedness between the generation ability of classifiers and GANs as they both utilize a classifier (discriminator) to guide image synthesis. We leave it to future works.

\textbf{AutoEncoder.} Our framework is different from autoencoders, which are typically for dimension reduction. The output image obtained by our method is only semantically related to the input but usually diverse in image contents. 

\textbf{Classifier Inversion.} Our framework is inspired by previous works on classifier inversion \citep{mahendran2015understanding,lopes2017data,yin2019dreaming,santurkar2019image}, which aims to synthesize training data by inverting a pretrained classifier. In contrast, we find that the filters from a trained classifier can be treated as a feature dictionary to build a universal and data-efficient image synthesizer. Besides, we find that this feature dictionary provides a natural way to tackle many sophisticated tasks such as unpaired image-to-image and feature similarity estimation. Compared with the robust classifier method \citep{santurkar2019image}, our framework does not require further network training.

\section{Visualization} \label{sec:visualization}

\subsection{Mean and Variance}

In the main paper, we have briefly discussed the role of mean and variance. Here we provide some experimental results to verify it. Figure \ref{sup:mean_var} shows three different coding schemas: "mean only", "variance only", and "mean+variance". As analyzed in the main paper, the mean statistic is sufficient enough to reconstruct the correct semantics. However, there are still some deficiencies in color and details. By introducing the variance, we can effectively improve the quality of synthesized images.

\begin{figure}[H]
  \centering
  \includegraphics[width=12cm]{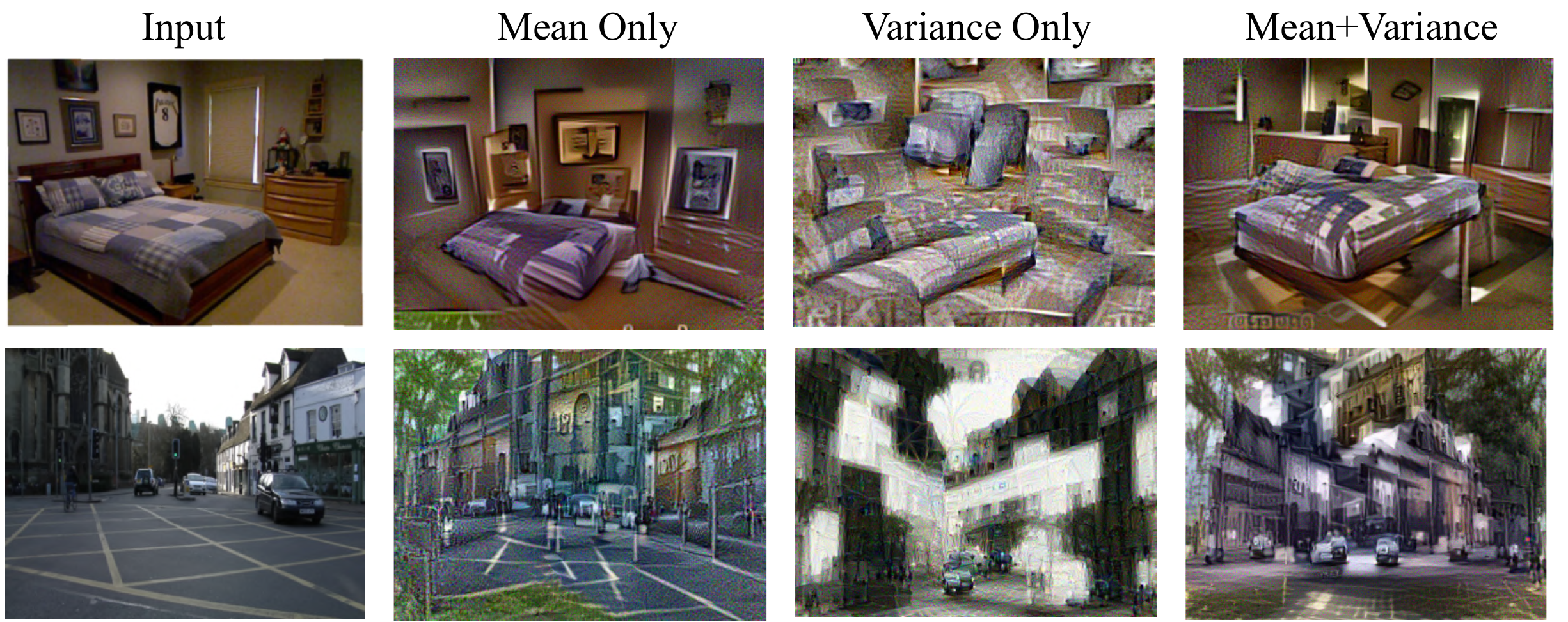}
  \caption{The role of mean and variance statistic in impression code. The mean statistic encodes image semantic and the variance encodes the diversity of image features. } \label{sup:mean_var}
\end{figure}

\subsection{More Experiment Results}

\begin{figure}[H]
  \centering
  \begin{subfigure}{13cm}
    \includegraphics[width=\textwidth]{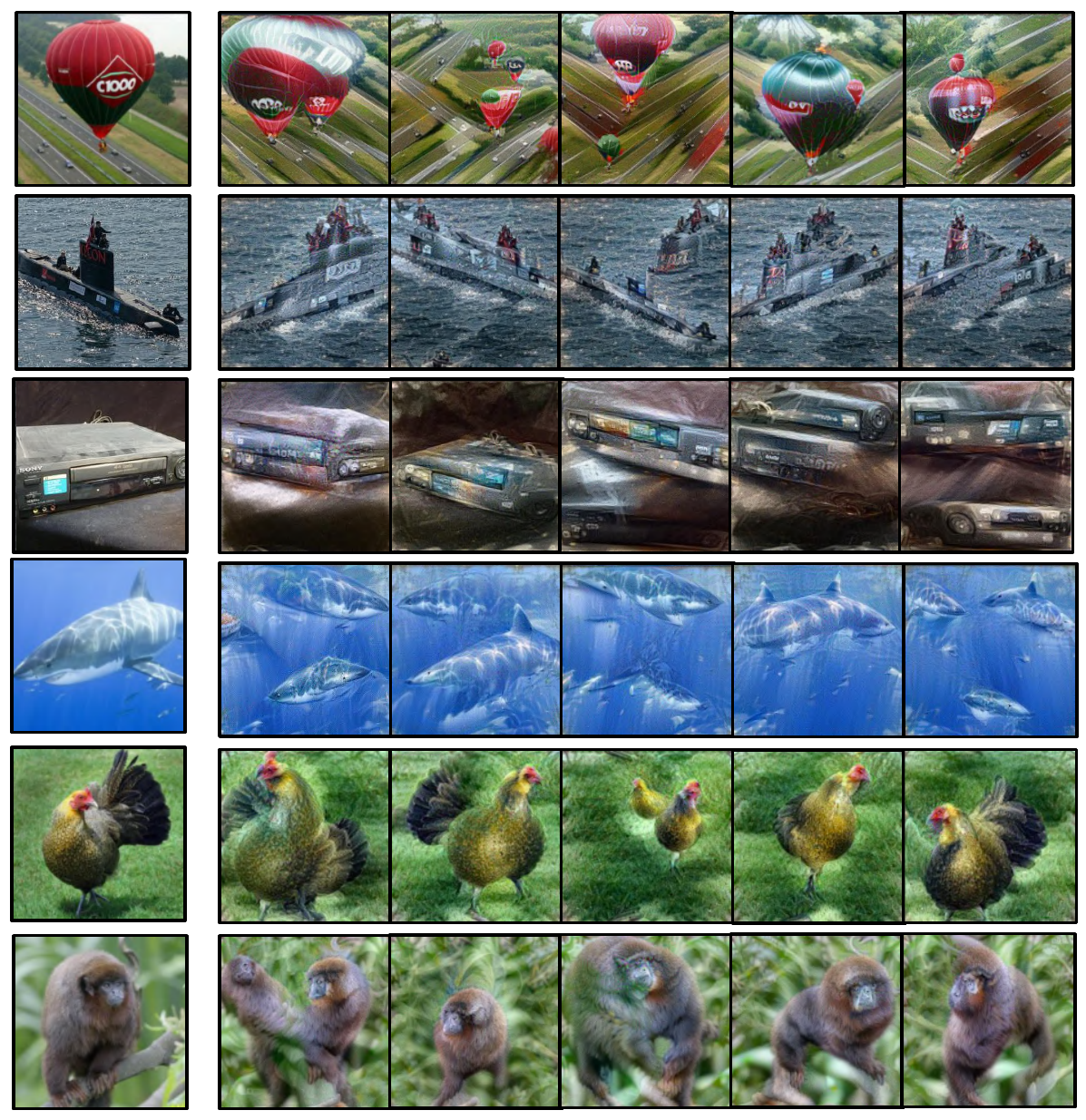}
    \caption{One-shot image synthesis on ImageNet.}
  \end{subfigure}
  
  \begin{subfigure}{13cm}
    \vspace{2mm}
    \includegraphics[width=\textwidth]{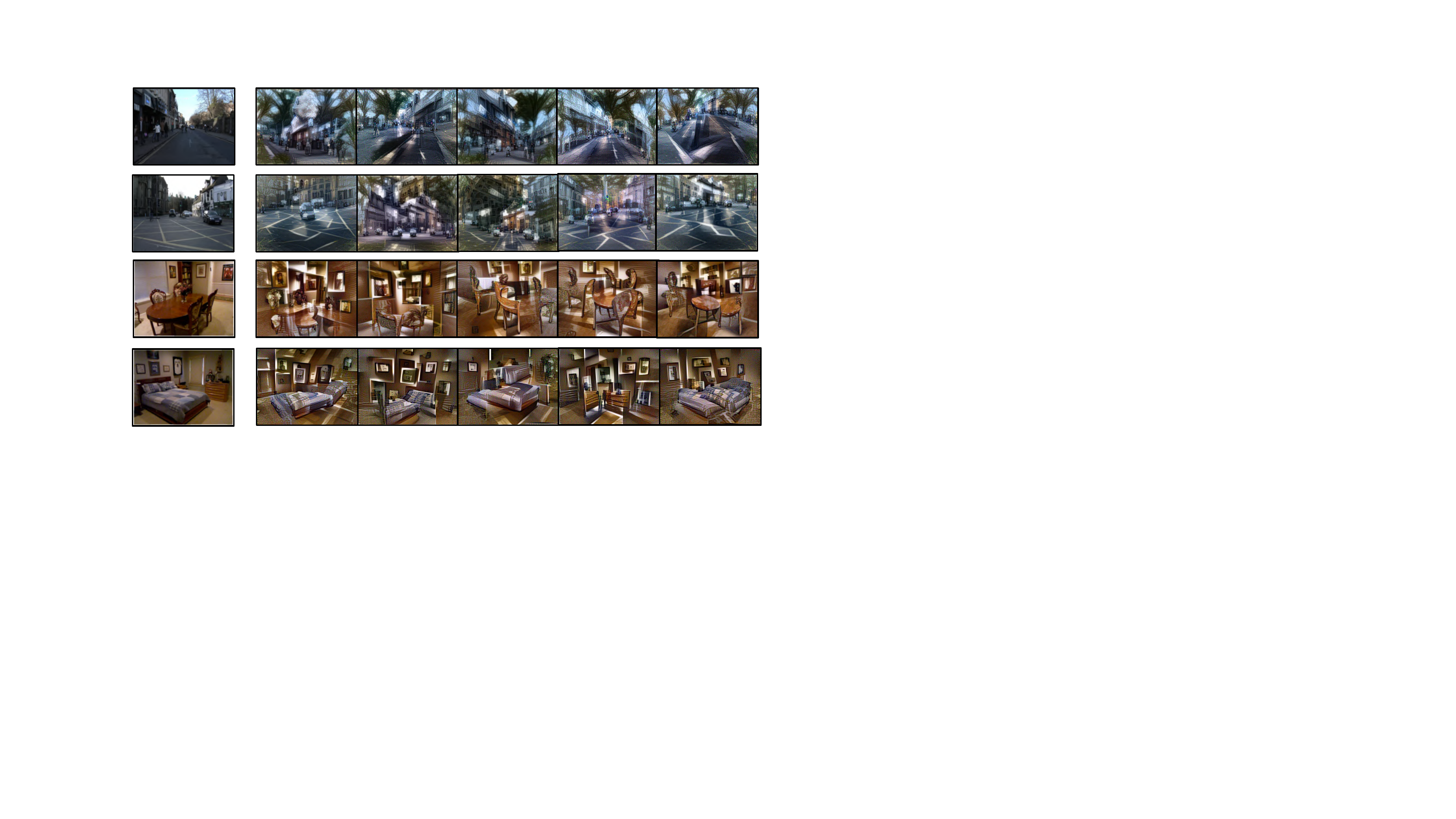}
    \caption{Universal image synthesis on CamVid and NYUv2.}
  \end{subfigure}
  \caption{Results of one-shot image synthesis. Synthesized images in each row are randomly sampled from the same batch.} \label{sup:synthesis}
\end{figure}

\begin{figure}[H]
  \centering
  \begin{subfigure}{13cm} 
    \includegraphics[width=\textwidth]{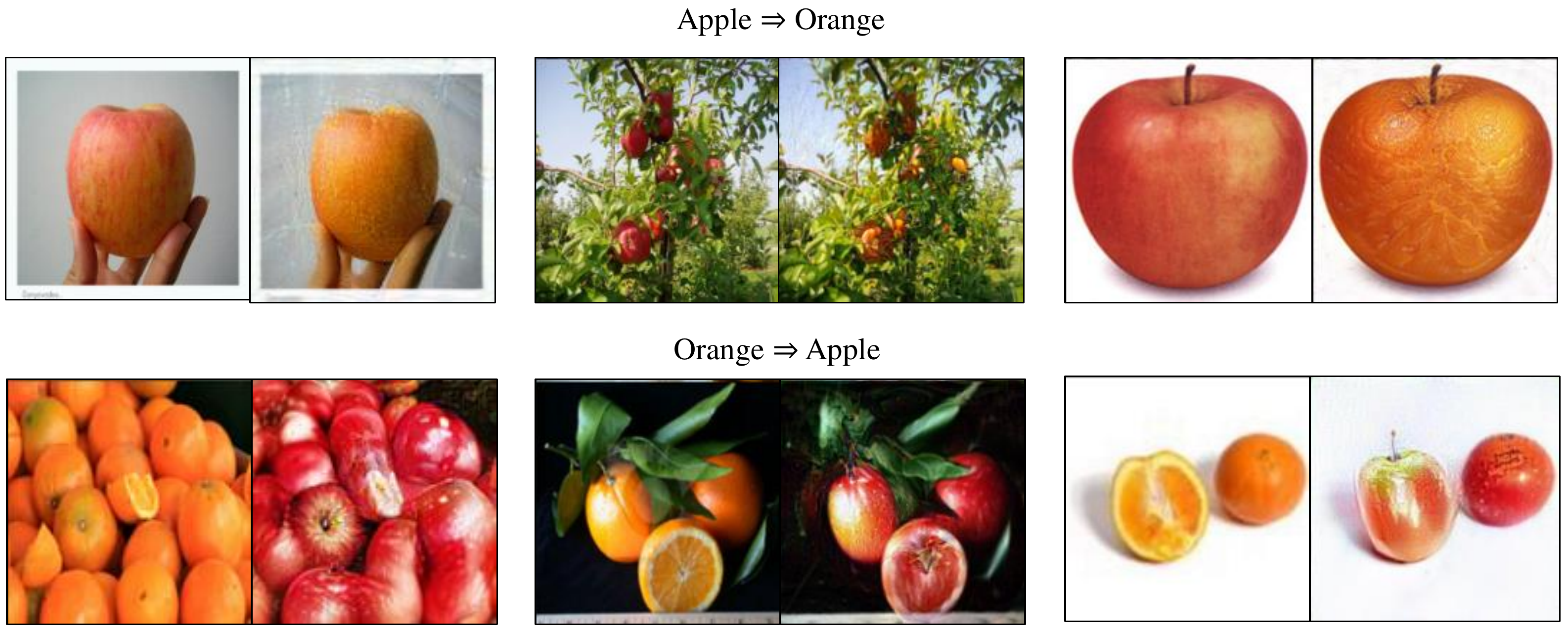}
    \includegraphics[width=\textwidth]{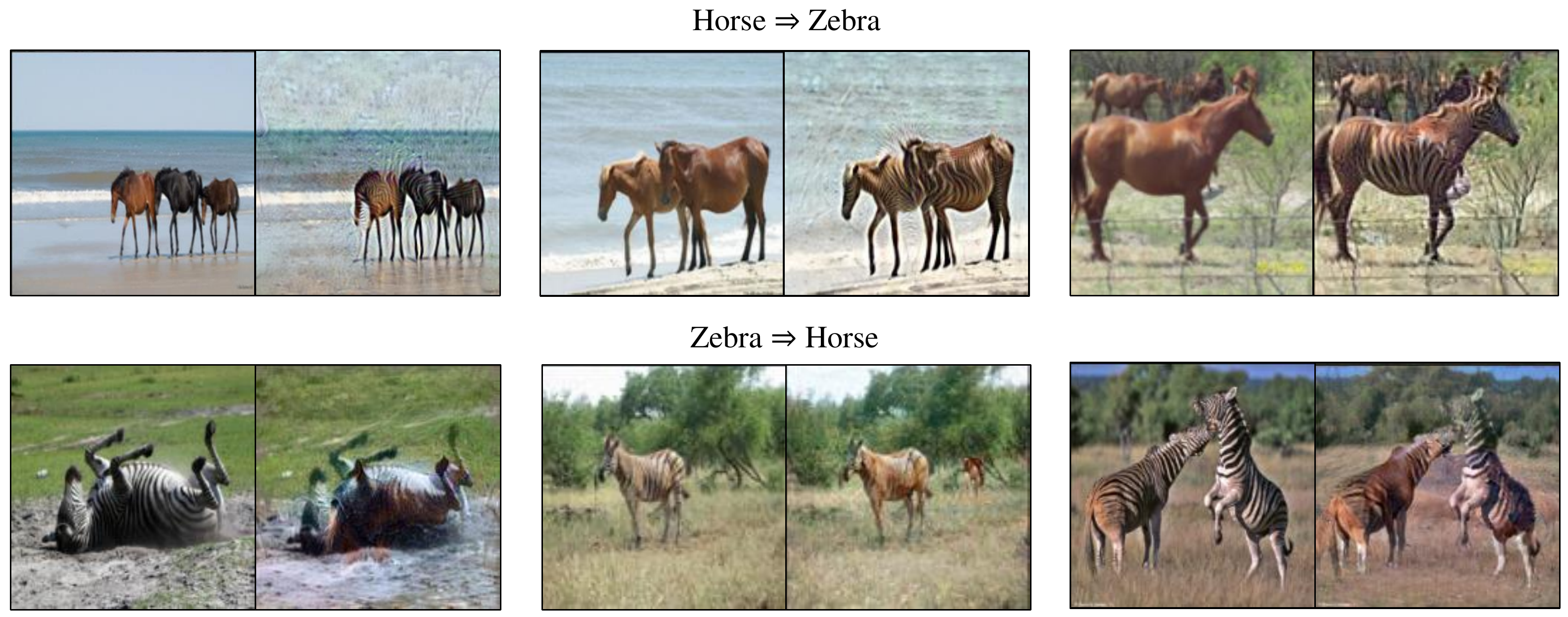}
    \includegraphics[width=\textwidth]{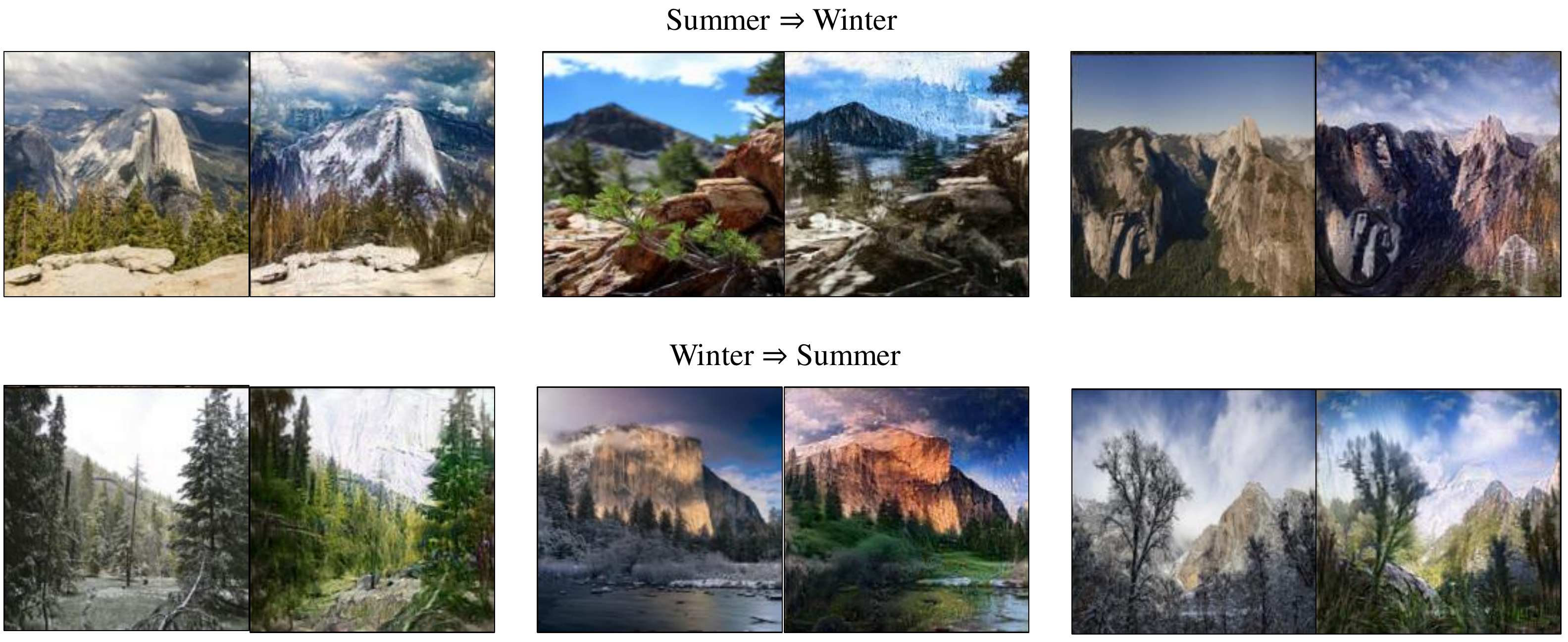}
    \includegraphics[width=\textwidth]{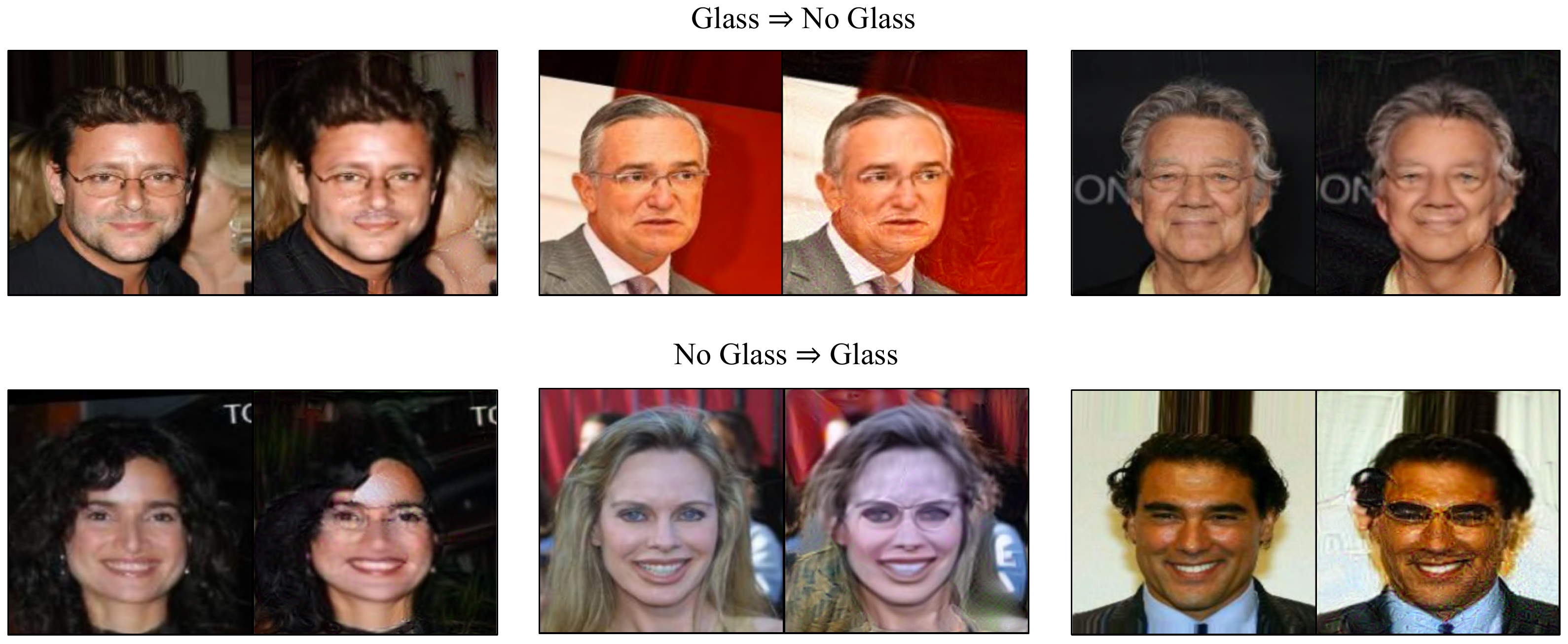}
  \end{subfigure} 
  \caption{Randomly sampled results from unpair image-to-image translation task. All images are translated by a pretrained ResNet-50 classifier.} \label{sup:translation}
\end{figure}

In this section, we provide more results from one-shot image synthesis and image-to-image translation. Figure \ref{sup:synthesis} illustrates the synthesized images on the classifier's training data (Imagenet  \citep{deng2009imagenet}) and universal data (CamVid \citep{brostow2008segmentation} and NYUv2 \citep{silberman2012indoor}). Synthesized images in each row are randomly sampled from the same batch. Figure \ref{sup:translation} provides results from different image translation tasks: $apple\leftrightarrow orange$, $horse\leftrightarrow zebra$, $summer\leftrightarrow winter$ and $glass\leftrightarrow no \; glass$. All results are obtained from an off-the-self ResNet-50 Classifier, and no network training is required.

\clearpage

\clearpage
\subsection{Failure Cases}

\begin{figure}[H]
  \centering
  \includegraphics[width=13cm]{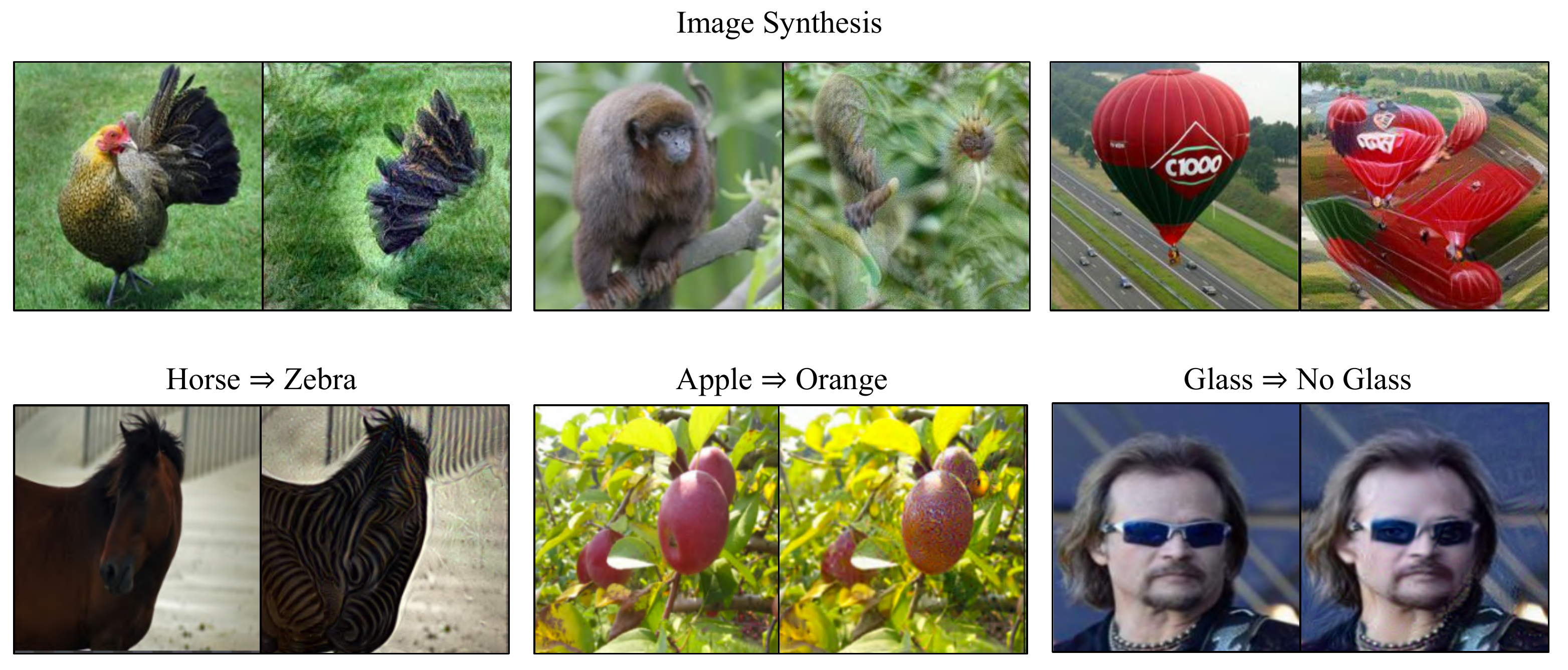}
  \caption{Some failure cases from image synthesis and translation.} \label{sup:mean_var}
\end{figure}

\subsection{Different Network Architectures}

\begin{figure}[H]
  \centering
  \includegraphics[width=13cm]{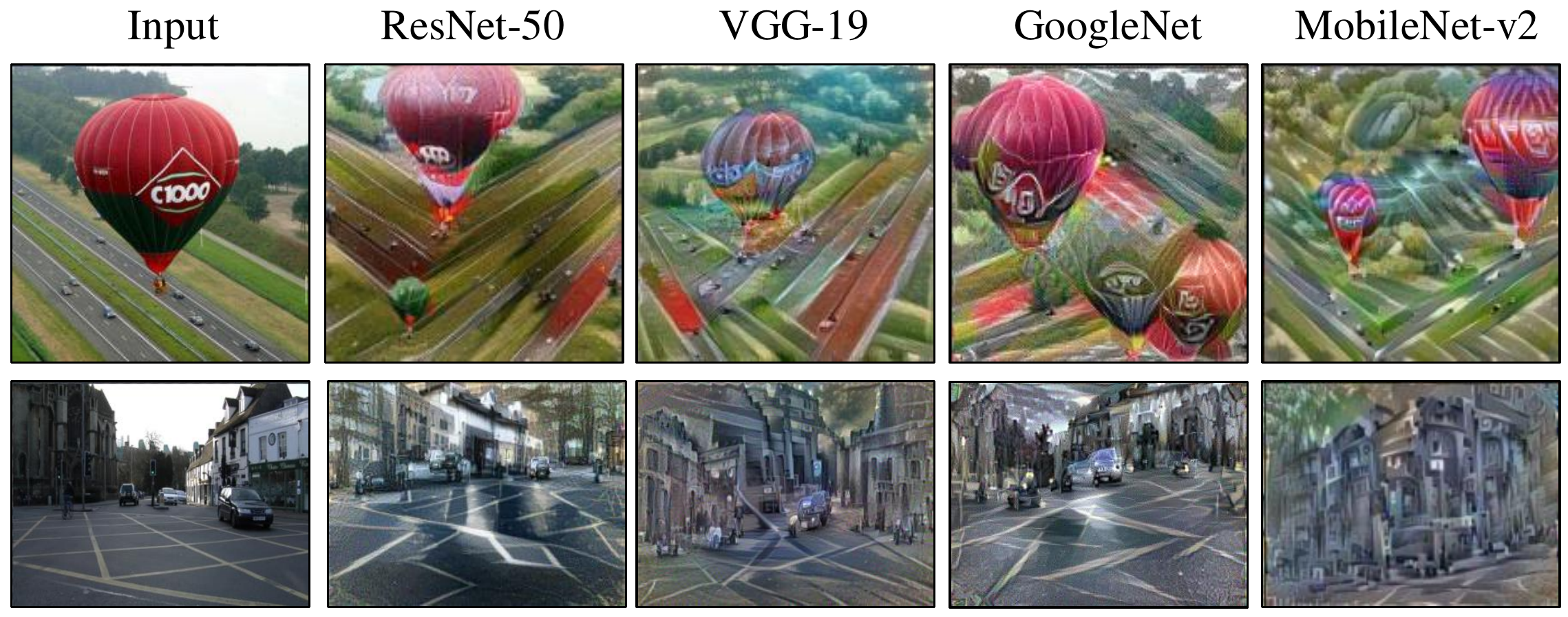}
  \caption{Synthesized images from different network architectures.} \label{sup:mean_var}
\end{figure}

\end{document}